\documentclass[11pt,a4paper]{article}
\usepackage[hyperref]{naaclhlt2019}
\usepackage{times}
\usepackage{latexsym}
\usepackage{amsmath}
\usepackage{graphicx}
\usepackage{comment}
\usepackage{booktabs}
\usepackage{enumitem}
\usepackage{rotating}
\setlist{label*=(\arabic*)}
\usepackage{url}
\usepackage{subcaption}
\usepackage{color} 
\usepackage{adjustbox}
\usepackage{multirow}
\usepackage{colortbl}
\usepackage{pifont}
\usepackage{multirow}
\usepackage{pifont}
\definecolor{cadmiumgreen}{rgb}{0.0, 0.42, 0.24}
\newcommand{\cmark}{\textcolor{cadmiumgreen}{\textbf{no}}}%
\newcommand{\xmark}{\textcolor{red}{\textbf{yes}}}
\newcolumntype{R}[2]{%
    >{\adjustbox{angle=#1,lap=\width-(#2)}\bgroup}%
    l%
    <{\egroup}%
}
\definecolor{LightCyan}{rgb}{0.88,1,1}

\aclfinalcopy 

\title{An Argument-Marker Model for Syntax-Agnostic Proto-Role Labeling}

\author{Juri Opitz {\normalfont and} Anette Frank \\
  Research Training Group AIPHES, \\
  Leibniz ScienceCampus ``Empirical Linguistics and Computational Language Modeling''\\
  Department for Computational Linguistics \\
  69120 Heidelberg \\
 {\tt $\lbrace$opitz,frank$\rbrace$@cl.uni-heidelberg.de} }

\date{}

\begin{document}
\maketitle
\begin{abstract}

Semantic proto-role labeling (SPRL) is an alternative to 
semantic role labeling (SRL) 
that
moves beyond 
a categorical definition of roles, following Dowty's feature-based view of \textit{proto-roles}. This theory determines \textit{agenthood} vs.\ \textit{patienthood} based on a participant's instantiation of more or less typical agent vs. patient properties, such as, for example, \textit{volition} in an event.
To perform SPRL, we develop an ensemble of hierarchical models with self-attention and concurrently learned predicate-argument-markers. Our method is competitive with the state-of-the art, overall outperforming previous work in two formulations of the task (multi-label and multi-variate Likert scale prediction).
In contrast to previous work, our results do not depend on gold argument heads derived from supplementary gold tree banks.
\end{abstract}

\section{Introduction}

Deciding on a linguistically sound, clearly defined and broadly applicable inventory of semantic roles is a long-standing issue in linguistic theory and natural language processing.
To alleviate issues found with classical thematic role inventories,
\citet{Dowty-1991} 
argued for replacing categorical roles 
with a feature-based, composite notion of semantic roles, introducing the \textit{theory of semantic proto-roles (SPR)}.
At its core, it proposes
two prominent, composite role types:
\textit{proto-agent} and \textit{proto-patient}.  
Proto-roles represent multi-faceted, possibly graded notions of \textit{agenthood} or \textit{patienthood}.
For example, consider the following sentence from Bram Stoker's \textit{Dracula} (1897):
\begin{enumerate}[resume]
  \item{\textit{He opened it} [the letter] \textit{and read it gravely.}\label{ex:1}}
\end{enumerate}
\begin{figure}
\footnotesize
\fbox{\begin{minipage}{23.5em}
\textbf{SR}:~~~ $\exists e
\big[open(e) \land agent(e, c^{*}) \land theme(e, l^{*})\big]$ \\
\textbf{SPR}: $\exists e
\big[open(e) \land volition(e, c^{*}) \land aware(e, c^{*}) \land sentient(e,c^{*}) \land manipulated(e, l^{*}) \land changes$-$state(e,l^{*}) \land ...\big]$
\end{minipage}}
\caption{Two different `Davidsionian' event analyses of \textit{open}. $c^{*}$ and $l^{*}$ refer to the \textit{count} and \textit{letter} entities.}
\label{ex:2}
\end{figure}

`Davidsonian' analyses based on SR and SPR  of the event \textit{open} are displayed in Figure \ref{ex:2}. The SPR analysis 
provides more detail about the event and the roles of the involved entities. 
Whether an argument is considered an
\textit{agent} or \textit{patient} follows 
from
the proto-typical properties the argument exhibits: e.g., \textit{being manipulated} is proto-typical for patient, while \textit{volition} is proto-typical for an agent. 
Hence, in both events of (1) the count is determined as agent, and the letter as patient.

Only recently two SPR data sets have been published. \citet{Q15-1034} developed a property-based proto-role annotation schema with 18 properties. 
One Amazon Mechanical Turk crowd worker (selected in a pilot annotation)
answered questions such as \textit{how likely is it
that the
argument 
mentioned with the verb changes location?}
on a 5-point Likert or responded \textit{inapplicable}. This dataset 
(news domain)
will henceforth 
be 
denoted by \textbf{SPR1}. Based on the experiences from the SPR1 annotation process,  \citet{white2016universal} published \textbf{SPR2} which follows a similar annotation schema. However, in contrast to SPR1, the new data set contains doubly annotated data from the web domain for 14 refined properties. 

Our work makes the following contributions:
In Section \S \ref{sec:2}, we provide an overview 
of
previous SPRL work and outline a common weakness: reliance on gold syntax trees or gold argument heads derived from them.
To alleviate this issue, we propose a span-based, hierarchical neural model (\S \ref{sec:mod}) which learns marker embeddings to highlight the predicate-argument structures of events. Our experiments (\S \ref{sec:exp}) show that our model, when combined in a simple voter ensemble, outperforms all previous works. A single model performs only slightly worse, albeit having weaker dependencies than previous methods. 
In our analysis, we (i) perform ablation experiments to analyze the contributions of different model components. (ii) we observe that the small SPR data size introduces a severe sensitivity to different random initializations of our neural model. We find that combining multiple models in a simple voter ensemble makes SPRL predictions not only slightly better but also significantly more robust. We share our code with the community and make it publicly available.\footnote{\url{https://gitlab.cl.uni-heidelberg.de/opitz/sprl}}

\section{Related Work}
\label{sec:2}
\paragraph{SPRL}\citet{DBLP:conf/aaai/TeichertPDG17} formulate the semantic role labeling task as a multi-label problem and develop
a conditional random field model (CRF). 
Given an argument phrase and a corresponding predicate, the model predicts which of the 18 properties hold.
Compared with a simple feature-based linear model by \citet{Q15-1034}, the CRF exhibits superior performance by more than 10 pp.\ macro F1.
Incorporating features derived from additional gold syntax improves the CRF performance significantly. For treating the task as a multi-label problem, the Likert classes $\{1,2,3\}$ and \textit{inapplicable} are collapsed to $-$ and Likert classes $\{4,5\}$ are mapped to $+$. Subsequent works, including ours, conform to this setup. 

\citet{DBLP:journals/corr/abs-1804-07976} are the first to 
treat SPRL as a multi-variate Likert scale regression problem.
They
develop a neural model whose predictions have good correlation with the values in the testing data of both SPR1 and SPR2. 
In the multi-label setting, their model compares favourably with \citet{DBLP:conf/aaai/TeichertPDG17} for most proto-role properties and establishes a new state-of-the-art. Pre-training the model in a machine translation setting helps on SPR1 but results in a performance drop on SPR2. 
The model takes
a sentence as input to
a Bi-LSTM \cite{hochreiter1997long} to produce a sequence of hidden states. \label{sec:relw}
The prediction is based on the hidden state corresponding to the head of the argument phrase, which
is determined by inspection of the gold syntax tree. 

Recently, \citet{tenney2018iclr} have demonstrated the capacities of contextualized word embeddings across a wide variety of tasks, including SPRL. However, for SPRL labeling they proceed similar to \citet{DBLP:journals/corr/abs-1804-07976} in the sense that they extract the gold heads of arguments in their dependency-based SPRL approach. Instead of using an LSTM to convert the input sentence to a sequence of vectors they make use of large language models such as ELMo \cite{peters2018deep} or BERT \citep{DBLP:journals/corr/abs-1810-04805}. The contextual vectors corresponding to predicate and the (gold) argument head are processed by a projection layer, self-attention pooling \citep{D17-1018} and a two-layer feed forward neural network with sigmoid output activation functions. To compare with \citet{DBLP:journals/corr/abs-1804-07976}, our basic model uses standard GloVe embeddings. When our model is fed with contextual embeddings a further observable performance gain can be achieved. 

To summarize, previous state-of-the-art SPRL systems suffer from a common problem: they are dependency-based and their results rely on gold argument heads. Our approach, in contrast, does not 
rely on any supplementary information from gold syntax trees. In fact, our marker model for SPRL is agnostic to any syntactic theory and acts solely on the basis of argument spans which we highlight with position markers. 

\paragraph{SRL} The task of automatically identifying predicate-argument structures and assigning roles to arguments was firstly investigated by \citet{gildea2002automatic}. Over the past years, SRL has witnessed a large surge in interest. Recently, very competitive end-to-end neural systems have been proposed \citep{he2018jointly,C18-1233,he2018syntax}. \citet{strubell2018linguistically} show that injection of syntax can help SRL models and \citet{DBLP:journals/corr/abs-1901-05280} bridge the gap between span-based and dependency-based SRL, achieving new state-of-the-art results both on the span based CoNLL data \citep{carreras2005introduction,pradhan2013towards} and the dependency-based CoNLL data \citep{surdeanu2008conll,hajivc2009conll}. A fully end-to-end S(P)RL system has to solve multiple sub-tasks: identification of predicate-argument structures, sense disambiguation of predicates and as the main and final step, labeling their argument phrases with roles. Up to the present, SPRL works (including ours) focus on the main task and assume the prior steps as complete.

Research into SPRL is still in its infancy, especially in comparison to SRL. One among many reasons may be the fact that, in contrast to semantic roles, semantic proto-roles are multi-dimensional. This introduces more complexity: given a predicate and an argument, the task is no more to predict \textit{a single} label (as in SRL), but a \textit{list} of multiple labels or even a multi-variate Likert scale. Another reason 
may be related to the available resources. The published SPR data sets comprise significantly fewer examples. The design of annotation guidelines and pilot studies with the aim of in-depth proto-role annotations is a hard task. In addition, the SPR data were created, at least to a certain extent, in an experimental manner: one of the goals of corpus creation was to explore possible SPR annotation protocols for humans. We hope that a side-effect of this paper is to spark more interest in SPR and SPRL. 

\section{Attentive Marker Model}
\label{sec:mod}
\begin{figure}
    \centering
    \includegraphics[scale=0.53]{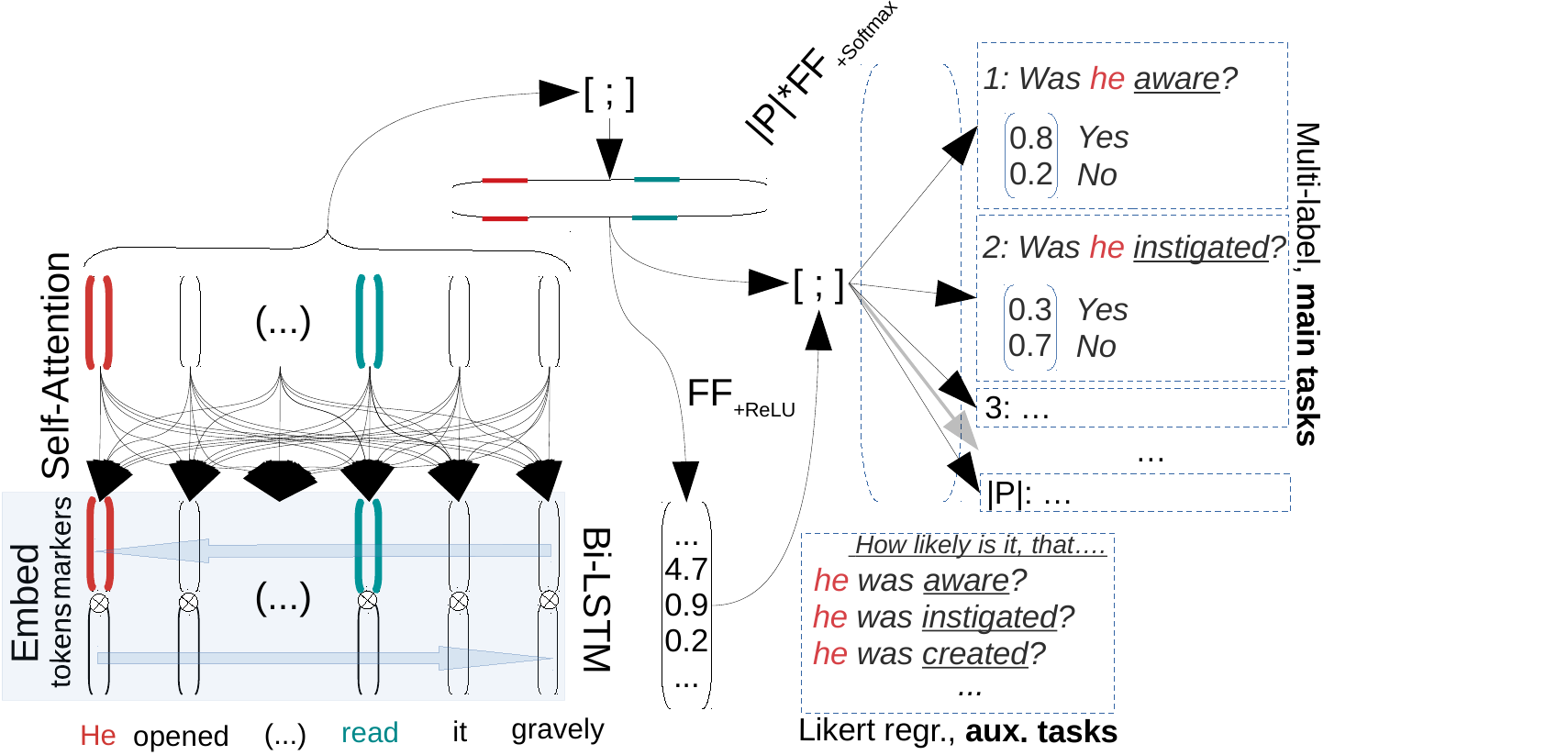}
    \caption{Model outline. \textit{Input}: (i) a sequence of vectors representing the words and (ii) a sequence of vectors which serve to highlight predicate and argument. \textit{Processing}: 1.\ element-wise multiplication of the two sequences ($\bigotimes$); 2.\ generation of hidden states with forward and backward Bi-LSTM reads (\includegraphics[angle=180,origin=c,scale=0.4]{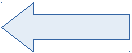} \&  \includegraphics[scale=0.4]{pdfresizer.pdf}); 3.\ self-attention mechanism builds a new sequence of hidden states by letting every hidden state attend to every other hidden state; 4.\ concatenation of the hidden states to generate a vector representation ($[;]$). \textit{Output}: (i) use vector representation to output Likert scale auxiliary predictions (FF$_{ReLU}$) and (ii) concatenate auxiliary predictions to the vector representation ($[;]$) to finally (iii) compute the multi-label predictions at the top level ($|P|\cdot$FF$_{Softmax}$; $P$: set of proto role properties).}
    \label{fig:mod}
\end{figure}
Since the work of \citet{DBLP:conf/aaai/TeichertPDG17}, the SPRL problem has been phrased as follows: given a \textit{(sentence, predicate, argument)} tuple, we need to predict for all possible argument properties from a given property-inventory whether they hold or not (regression: \textit{how likely are they to hold?}).

Following previous work \citep{DBLP:journals/corr/abs-1804-07976}, the backbone of our model is a Bi-LSTM. To ensure further comparability, pretrained 300 dimensional GloVe embeddings \citep{pennington2014glove} are used for building the input sequence $(e_1,...,e_T)$. In contrast to \citet{DBLP:journals/corr/abs-1804-07976}, we multiply a sequence of \textit{marker embeddings} $(m_1,...,m_T)$ element-wise with the sequence of word vectors: $(e_1\cdot m_1,...,e_T\cdot m_T)$ ($\bigotimes$, Figure \ref{fig:mod}). 
We distinguish three different
marker embeddings 
that
indicate the position of the argument in question ($\color{red}{red}$, Figure \ref{fig:mod}), the predicate  ($\color{cadmiumgreen}{green}$, Figure \ref{fig:mod}) and 
remaining parts of the sentence. This is to some extent similar to \citet{he17} who learn two predicate indicator embeddings which are concatenated to the input vectors and serve the purpose of showing the model whether a token is the predicate or not. However, in SPRL we are also provided with the argument phrase. We will see in the ablation experiments that it is paramount to learn a dedicated embedding. Embedding multiplication instead of concatenation has the advantage of fewer LSTM parameters (smaller input dimension). Besides, it provides the model with the option to learn large coefficients of the word vector dimensions of predicate and argument vectors. This should immediately draw the model's attentiveness to the argument and predicate phrases which now are accentuated.

The sequence of marked embeddings is further processed by a Bi-LSTM in order to obtain a sequence of hidden states $S=(s_1,...,s_T)$. In Figure \ref{fig:mod}, forward and backward LSTM reads are indicated by  \includegraphics[angle=180,origin=c,scale=0.4]{pdfresizer.pdf} and  \includegraphics[scale=0.4]{pdfresizer.pdf}.

From there, we take intuitions from \citet{zheng2018opentag} and compute the next sequence of vectors by letting every hidden state attend to every other hidden state, which is expressed by the following formulas:
\begin{align*}
  h_{t,t'} &= \tanh (QS_t + KS_{t'} + \beta)\\
  e_{t,t'} &= \sigma(v^Th_{t,t'} + \alpha)\\
  a_{t}  &= softmax (e_{t})\\
  z_{t} &=\sum_{t'}a_{t,t'}\cdot s_{t'}
\end{align*}
\begin{figure}
    \centering
    \includegraphics[width=\linewidth]{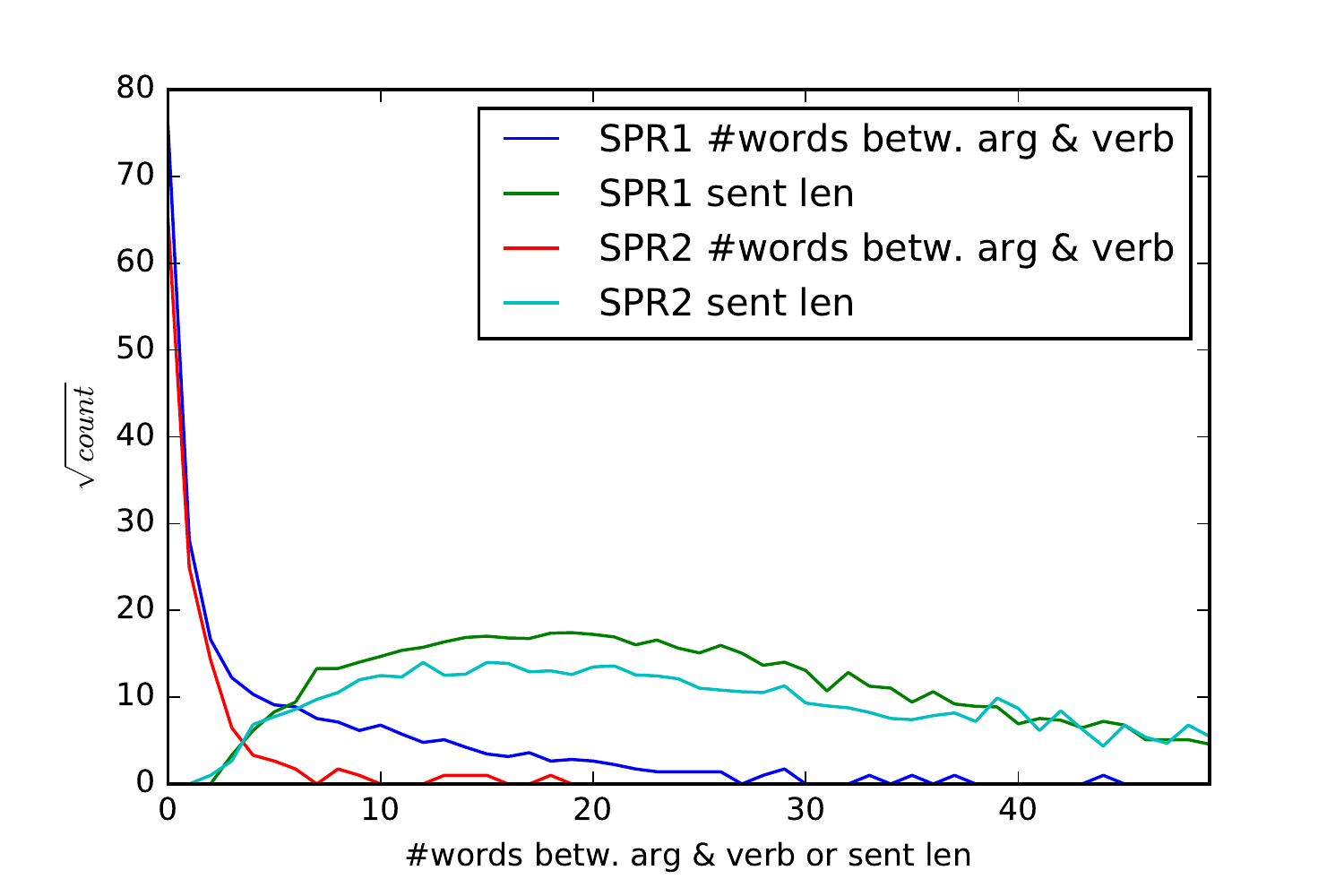}
    \caption{Distribution of the number of words between argument and verb (distance relationship) and sentence lengths in the data sets SPR1 and SPR2.}
    \label{fig:datastats}
\end{figure}
$Q, K$ are weight matrices, $\beta$ is a bias vector, $\alpha$ is a bias scalar and $v$ a weight vector. Letting every hidden state attend to every other hidden state gives the model freedom in computing the argument-predicate composition. This is desirable, since arguments and predicates frequently are in long-distance relationships. For example, in Figure \ref{fig:datastats} we see that in the SPR1 data predicates and arguments often lie more than 10 words apart and a non-negligible amount of cases consists of distances of more than 20 words.

We proceed by concatenation, $\mathbf{z}=[z_1;... ; z_T]$,
and compute intermediate outputs approximating the property-likelihood Likert scales. This is achieved with
weight matrix $A$ and $ReLU$ activation functions ($FF_{+ReLU}$, Figure \ref{fig:mod}):

\begin{equation}
    a=ReLU(A\mathbf{z}).
\end{equation}

To perform
multi-label prediction with $|P|$ possible labels 
we use 
$[a;\mathbf{z}]$
for computing the final decisions with $2|P|$ output neurons and $|P|$ separate weight matrices ($|P|^{*}FF_{+softmax}$, Figure \ref{fig:mod}), one for each property $p \in P$:

\begin{equation}
    o_{p}=softmax(W^{p}[a;\mathbf{z}]).
\end{equation}

For example, given the 18 proto-role properties contained in SPR1, we learn $|P|=18$ weight matrices and use the Softmax functions to produce 18 vectors of dimension 2 as outputs. The first dimension $o_{p,0}$ represents the predicted probability that property $p$ does \textit{not apply} ($o_{p,1}$: probability for \textit{$p$ applies}). For the regression task, we reduce the number of output neurons from $2|P|$ to $|P|$ and use $ReLU$ activation functions instead. We 
hypothesize
that the hierarchical structure can support the model in making predictions on the top layer. E.g., if the argument is predicted to be most likely \textit{not sentient} and very likely to be \textit{manipulated}, the model may be less tempted to predict an \textit{awareness} label at the top layer. The auxiliary loss for any datum is given as the mean square error over the auxiliary output neurons:
\begin{equation}
    \ell' = \frac{\lambda'}{|P|}\sum_{p \in P} (a^\star_{p}-a_{p})^2
\end{equation}
In case of the multi-label formulation, our main loss for an example is the average cross entropy loss over every property:
\begin{equation}
    \ell = -\frac{\lambda}{|P|}\sum_{p \in P}(o^\star_{p,1}\log{o_{p,1}} + o^\star_{p,2}\log{o_{p,2}}),
\end{equation}
where $o^\star_{p,0} = \mathbf{I(\neg p)}$ and $o^\star_{p,1} = \mathbf{I(p)}$ i.e.\ the gold label indicator. 
\section{Experiments}
\label{sec:exp}

\paragraph{Data} We use the same data setup and split as \citet{DBLP:conf/aaai/TeichertPDG17,DBLP:journals/corr/abs-1804-07976,tenney2018iclr}.  
 For determining the gold labels, we also conform to prior works and (i) collapse classes in the multi-label setup from $\{NA,1,2,3\}$ and $\{4,5\}$ to classes `$-$' and `$+$' and (ii) treat $NA$ as 1 in the Likert regression formulation. For doubly annotated data (SPR2), the Likert scores are averaged; in the multi-label setup we consider values $\geq 4$ as `$+$' and map lesser scores to `$-$'. More data and pre-processing details are described in the Supplement \S \ref{sec:notes}.

\paragraph{Baselines} As baselines we present the results from previous systems: the state-of-the-art by \citet{DBLP:journals/corr/abs-1804-07976} is denoted in our tables as \textbf{RUD'18}, the linear feature-based classifier by \citet{Q15-1034} as \textbf{REI'15} and the CRF developed by \citet{DBLP:conf/aaai/TeichertPDG17} as \textbf{TEI'17}. Like previous works, we use macro F1 as the global performance metric in the multi-label scenario and macro-averaged Pearson's $\rho$ (arithmetic mean over the correlation coefficients for each property, details can be found in the Supplement \ref{sec:notes}) We refer to the system results as reported by \citet{DBLP:journals/corr/abs-1804-07976}. The most recent work, which evaluates large language models on a variety of tasks including SPRL, is denoted by \textbf{TEN'19} \citep{tenney2018iclr}. In this case, we present the micro F1 results as reported in their paper. 

\paragraph{Model instantiation}
We introduce four main models: (i) \textbf{Marker}: our basic, span-based single-model system. For (ii) \textbf{MarkerE}, we fit an ensemble of 50 \textbf{Marker}s with different random seeds. Computationally, training 50 neural models in this task is completely feasible since neither SPR1 nor SPR2 contain more than 10,000 training examples (parallelized training took approximately 2 hours). The ensemble predicts unseen testing data by combining the models' decisions in a simple majority vote when performing multi-label prediction or, when in the regression setup, by computing the mean of the output scores (for every property).   We also introduce (iii) \textbf{MarkerB}, and (iv) \textbf{MarkerEB}. These two systems differ in only one aspect from the previously mentioned models: instead of GloVe word vectors, we feed contextual vectors extracted from the BERT model \citep{DBLP:journals/corr/abs-1810-04805}. More precisely, we use the transformer model \textit{BERT-base-uncased}\footnote{\url{https://github.com/google-research/bert}} and sum the inferred activations over the last four layers. The resulting vectors are concatenated to GloVe vectors and then processed by the Bi-LSTM.

We fit all models with gradient descent and apply early stopping on the development data (maximum average Pearson's $\rho$ for multi-variate Likert regression, maximum macro F1 for the multi-label task). Further hyper parameter choices and details about the training are listed in Appendix \S \ref{sec:hyp}.

\subsection{Multi-Label Prediction Results}
\begin{table*}[ht]
\centering
\scalebox{0.63}{\begin{tabular}{l|rrr|rrrr|||r|rrrr|} 
\toprule
&\multicolumn{7}{c|||}{multi-label (ML), F1 score}&\multicolumn{5}{c}{regression\ (LR), $\rho$}\\
\cmidrule{2-13}
&\multicolumn{3}{c|}{previous works}&\multicolumn{4}{c|||}{ours}&&\multicolumn{4}{c}{ours}\\
property & REI'15 & TEI'17 & RUD'18 & MarkerE &Marker&MarkerEB&MarkerB& RUD'18 & MarkerE & Marker & MarkerEB & MarkerB \\
\midrule 
awareness & 68.8 & 87.3 & 89.9 & 89.4&88.1&\textit{\textbf{93.4}}&\textbf{91.6}& 0.897  & 0.880&0.868 &\textbf{0.922}&\textbf{0.915}\\
\rowcolor{LightCyan}chg location & 6.6 & 35.6&45.7 & \textbf{50.5}&\textbf{53.2}&\textit{\textbf{64.6}}&\textbf{57.1}&0.702  & \textbf{0.768}&\textbf{0.744}&\textbf{0.835}&\textbf{0.814}\\
chg state & 54.6&66.1&71.0 & 66.6&68.2&\textit{\textbf{71.5}}&66.1&0.604  & \textbf{0.651}&\textbf{0.621}&\textbf{0.701}&\textbf{0.653}\\
\rowcolor{LightCyan}chg possession & 0.0 & 38.8&58.0 & 48.4&55.5&\textbf{55.7}&\textit{\textbf{59.5}}&0.640  & 0.609&0.576&\textit{\textbf{0.652}}&0.582\\
created & 0.0&44.4& 39.7  & 34.8&46.5&\textbf{52.2}&\textit{\textbf{56.1}}&0.549 & \textbf{0.593}&0.498&\textit{\textbf{0.669}}&\textbf{0.614}\\
\rowcolor{LightCyan}destroyed &17.1&0.0&24.2  & \textit{\textbf{26.6}}&19.9&14.3&22.2&0.346 & \textbf{0.368}&0.220&\textit{\textbf{0.412}}&\textbf{0.347}\\
existed after & 82.3 & 87.5& 85.9  & 87.0&82.9&\textit{\textbf{88.5}}&84.8&0.619  & 0.612&0.571&\textit{\textbf{0.695}}&\textbf{0.669}\\
\rowcolor{LightCyan}existed before & 79.5& 84.8&85.1  & \textbf{87.4}&\textbf{85.4}&\textit{\textbf{90.1}}&\textbf{85.8}&0.710  & 0.704&0.668&\textit{\textbf{0.781}}&\textbf{0.741}\\
existed during & 93.1 & 95.1 &95.0 & \textbf{95.2}&94.3&\textit{\textbf{95.8}}&94.0& 0.673   & \textbf{0.675}&0.626&\textit{\textbf{0.732}}&\textbf{0.714}\\
\rowcolor{LightCyan}exists as physical & 64.8 &76.4& 82.7 & \textbf{83.7}&80.6&\textit{\textbf{88.1}}&\textbf{85.8}& 0.834 & 0.807&0.777&\textit{\textbf{0.871}}&\textbf{0.856}\\
instigated & 76.7 & 85.6 & 88.6 &  86.6 &86.4&\textit{\textbf{88.9}}&87.6&  0.858 & 0.856&0.842&\textit{\textbf{0.879}}&\textbf{0.860}\\
\rowcolor{LightCyan}location & 0.0 & 18.5&53.8  & \textbf{72.9}&\textbf{69.8}&\textit{\textbf{78.9}}&\textbf{76.1}&0.619  & \textbf{0.755} &\textbf{0.742}&\textit{\textbf{0.849}}&\textbf{0.820}\\
makes physical contact & 21.5 & 40.7& 47.2  & 45.7&32.3&\textit{\textbf{67.0}}&\textbf{61.1}&0.741 & 0.716&0.671&\textit{\textbf{0.801}}&\textbf{0.772}\\
\rowcolor{LightCyan}manipulated & 72.1 & 86.0 & 86.8 & \textbf{86.9}&86.7&\textit{\textbf{89.6}}&86.5&0.737 & \textbf{0.738}&0.705&\textit{\textbf{0.774}}&\textbf{0.751}\\
pred changed arg & 54.0&67.8&70.7  & 68.1&67.6&\textit{\textbf{72.8}}&66.1&0.592  & \textbf{0.621}  &0.579&\textit{\textbf{0.714}}&\textbf{0.664}\\
\rowcolor{LightCyan}sentient & 42.0 & 85.6& 90.6 & 89.5&88.3&\textit{\textbf{95.4}}&\textbf{92.2}&0.925  & 0.904&0.887&\textit{\textbf{0.959}}&\textbf{0.951}\\
stationary & 13.3 & 21.4 & 47.4 & 41.0&25.0&\textbf{50.5}&\textit{\textbf{54.5}}&0.711 & 0.666&0.654&\textit{\textbf{0.771}}&\textbf{0.739}\\
\rowcolor{LightCyan}volitional & 69.8 & 86.4 & 88.1 & 88.3&86.7&\textit{\textbf{91.6}}&\textbf{90.1}&0.882 & 0.873&0.863&\textit{\textbf{0.911}}&\textbf{0.903}\\ 
\midrule
micro &71.0& 81.7&83.3 &\textbf{83.6}&82.0&\textit{\textbf{86.8}}&\textbf{83.5}&-&-&-&-&- \\
macro &55.4&65.9&71.1&\textbf{72.1}&69.3&\textit{\textbf{77.5}}&\textbf{73.8}&0.706&\textbf{0.711}&0.673&\textit{\textbf{0.774}}&\textbf{0.743}\\
\bottomrule
\end{tabular}}
\caption{SPR1 results.  
\textbf{bold}: better than all previous work; \textit{\textbf{bold}}: overall best.
}
\label{tab:comprudtei}
\end{table*}
\begin{table*}
\centering
\scalebox{0.68}{\begin{tabular}{l|rrr|rrrr|||r|rrrr|} 
\toprule
&\multicolumn{7}{c|||}{multi-label (ML), F1 score}&\multicolumn{5}{c}{regression\ (LR), $\rho$}\\
\cmidrule{2-13}
&\multicolumn{3}{c}{baselines}&\multicolumn{4}{c|||}{ours}&&\multicolumn{4}{c}{ours}\\
property & maj&ran & const & MarkerE &Marker&MarkerEB&MarkerB& RUD'18 & MarkerE & Marker & MarkerEB & MarkerB   \\ 
\midrule 
awareness & 0.0 &48.9 & 67.1 & \textbf{92.7}&\textbf{92.3}&\textit{\textbf{94.0}}&\textbf{91.1}&0.879  & \textbf{0.882}&0.868&\textit{\textbf{0.902}}&0.878\\
\rowcolor{LightCyan}chg location&0.0  & 12.0&21.7 & \textbf{28.6}&\textbf{35.1}&\textit{\textbf{38.0}}&18.2&0.492  & \textbf{0.517}&0.476&\textit{\textbf{0.563}}&\textbf{0.507}\\
chg possession  &0.0& 5.5 & 6.6 & \textbf{33.3}&\textbf{33.3}&\textbf{35.6}&\textit{\textbf{41.1}}&0.488  &\textbf{ 0.520}&0.483&\textit{\textbf{0.549}}&\textbf{0.509}\\
\rowcolor{LightCyan}chg state&0.0 &19.5&31.3 & 29.7&27.1&\textbf{41.4}&\textit{\textbf{45.2}}&0.352  & 0.351&0.275&\textit{\textbf{0.444}}&\textbf{0.369}\\
chg state continuous &0.0&9.2&21.7 & \textbf{25.3}&19.8&\textbf{26.8}&\textit{\textbf{30.4}}&0.352  & \textbf{0.396}&0.321&\textit{\textbf{0.483}}&\textbf{0.423}\\
\rowcolor{LightCyan}existed after &94.1 & 86.1& 94.1  & 94.0&92.4&94.0&\textit{\textbf{94.5}}&0.478  & 0.469&0.403&\textit{\textbf{0.507}}&0.476\\
existed before & 89.5& 80.0&89.5  & \textbf{91.0}&\textbf{90.5}&\textit{\textbf{92.0}}&\textbf{89.8}&0.616  & \textbf{0.645}&0.605&\textit{\textbf{0.690}}&\textbf{0.664}\\
\rowcolor{LightCyan}existed during & 98.0 &96.2 &97.0 & 98.0&97.8&\textit{\textbf{98.1}}&\textit{\textbf{98.1}}& 0.358   & \textit{\textbf{0.374}}&0.280&0.354&0.301\\
instigated &0.0 & 48.9 & 70.5 &  \textbf{77.9} &\textbf{78.0} &\textit{\textbf{78.9}}&\textbf{78.7}& 0.590 & 0.582&0.540&\textit{\textbf{0.603}}&\textbf{0.599}\\
\rowcolor{LightCyan}partitive &0.0&10.4&\textit{\textbf{24.2}} & 2.5&16.5&9.2&2.4&0.359  & 0.283&0.213&\textit{\textbf{0.374}}&0.330\\
sentient & 0.0 &47.6& 44.3 & \textbf{91.9}&\textbf{91.6}&\textit{\textbf{93.7}}&\textbf{92.0}&0.880  & 0.874&0.859&\textit{\textbf{0.892}}&0.872\\
\rowcolor{LightCyan}volitional &0.0 & 39.1 & 61.8 & \textbf{88.1}&\textbf{87.2}&\textit{\textbf{89.7}}&\textbf{88.5}&0.841 & 0.839&0.825&\textit{\textbf{0.870}}&\textbf{0.854}\\ 
was for benefit&0.0 &31.6&48.8 & \textbf{61.1}&\textbf{59.2}&\textbf{60.2}&\textit{\textbf{63.4}}&0.578  & \textbf{0.580}&0.525&\textit{\textbf{0.598}}&0.569\\
\rowcolor{LightCyan}was used&79.3 &66.1&79.3 & 77.9&78.0&77.6&\textit{\textbf{79.9}}&0.203  & 0.173&0.093&\textit{\textbf{0.288}}&\textbf{0.264}\\
\midrule
micro &65.0& 62.9&61.4 &\textbf{84.0}&\textbf{83.4}&\textit{\textbf{84.9}}&\textbf{83.9}&-&-&-&-&- \\
macro&25.9 & 43.2&61.4&\textbf{70.9}&\textbf{69.7}&\textbf{69.9}&\textbf{67.5}&0.534&\textbf{0.535}&0.483&\textit{\textbf{0.580}}&\textbf{0.544}\\
\bottomrule
\end{tabular}}
\caption{SPR2 results. \textbf{bold}:  better
than previous work and/or baselines;
\textit{\textbf{bold}}: overall best.
}
\label{tab:comprudteispr2}
\end{table*}

\begin{table*}[t]
\newcommand*\rot{\rotatebox{90}}
\newcommand*\OK{\ding{51}}
    \centering
    \scalebox{0.64}{
    \begin{tabular}{@{}ll|r|llll|rrrr|rr@{}}
            &&&&& && \multicolumn{4}{c}{macro result}& \multicolumn{2}{c}{micro result}\\ 
            \cmidrule{3-13}
            && & \multicolumn{4}{c|}{used by system}  & \multicolumn{2}{c}{SPR1} & \multicolumn{2}{c}{SPR2}&SPR1&SPR2\\
          &Method   &\textit{span} or \textit{dep.} &gold head & STL & ensembling &  C-embeddings&ML & LR & ML & LR&ML&ML\\
            \midrule
        \multirow{10}*{\rotatebox{90}{ours~~~~~~~~~~~~~~~~~previous}}  &REI'15  & span&\cmark&\cmark & \cmark & \cmark&  55.4 &- & -&-&71.0&-\\
        &TEI'17   & dependency &\xmark~(full parse)&\cmark & \cmark & \cmark& 65.9 &- & -&-&81.7&-\\
        &RUD'18  & dependency&\xmark&\cmark & \cmark & \cmark&  69.3 &0.697 & -&0.534 & 82.2&-\\
        &RUD'18$_{(+MT~pretrain)}$  & dependency&\xmark&\xmark & \cmark & \cmark&  71.1 &0.706 & -&0.521 & 83.3&-\\
        &TEN'19  & dependency&\xmark&\cmark & \cmark & \xmark~(BERT concat)&  - &- & -&- & 84.7 & 83.0\\
        &TEN'19  & dependency&\xmark&\cmark & \cmark & \xmark~(BERT lin.\ comb.)&  - &- & -&- & 86.1$^\dagger$ & 83.8\\
        \midrule
        &Marker  & span&\cmark& \cmark & \cmark & \cmark &69.3& 0.682 &67.9$^\dagger$ &0.483 &  80.8 &81.1 \\
        &MarkerE  & span&\cmark &\cmark & \xmark & \cmark &\underline{72.1}& \underline{0.711} &\textbf{\underline{70.9}} &\underline{0.535} & \underline{83.6} &\underline{84.0}\\
         &MarkerB& span&\cmark& \cmark & \cmark & \xmark~(BERT sum) &73.5$^\dagger$& 0.741$^\dagger$ &67.5 &0.544$^\dagger$ &84.9 &83.9$^\dagger$ \\
       &MarkerEB
        & span&\cmark& \cmark & \xmark & \xmark~(BERT sum) &\textbf{77.5}& \textbf{0.774} &69.9 &\textbf{0.580} & \textbf{86.8} & \textbf{84.9}\\
    \end{tabular}}
    \caption{Main results, system properties and requirements of SPRL systems. Overall best system is marked in \textbf{bold}, best system using GloVe is \underline{underlined}, best single-model system is marked by $\dagger$.  STL: supervised transfer-learning (e.g., RUD'18: pre-training on MT task). C-embeddings: contextual word embeddings (BERT-base). ML: multi-label prediction; LR: multi-variate Likert regression. $BERT~concat$: last four BERT layers are concatenated. $BERT~lin.~comb.$: optimized linear combination of last four BERT layers (our BERT based models sum the last four layers). The Table is further discussed in the Section \textit{Discussion} \S \ref{par:discuss}.
     }
    \label{tab:my_label}
\end{table*}

\paragraph{News texts} The results on newspaper data (SPR1) are displayed in Table \ref{tab:comprudtei} (left-hand side). Our basic ensemble (\textbf{MarkerE}) improves massively in the property \textit{location}\footnote{i.e.\ \textit{does the argument describe the location of the event?}} (+19.1 pp.\ F1). A significant loss is experienced in the property \textit{changes possession} (-9.6).  
Overall, our ensemble method outperforms all prior works (REI'15: +17.7 pp.\ macro F1; TEI'17: +6.2, RUD'18: +1.0). Our ensemble method provided with additional contextual word embeddings (\textbf{MarkerEB}) yields another large performance gain. The old state-of-the-art, RUD'18, is surpassed by more than 6.0 pp.\ macro F1 (a relative improvement of 8.6\%). With regard to some properties, the contextual embeddings provide massive performance gains over our basic \textbf{MarkerE}: \textit{stationary} (+9.5 pp.\ F1), \textit{makes physical contact}  (+21.3), \textit{change of location} (+14.1) and \textit{created} (+17.3). The only loss is incurred for the property which asks if an argument is \textit{destroyed} (-12.3). This specific property appears to be difficult to predict for all models. The best score in this property is achieved by \textbf{MarkerE} with only 26.6 F1.
 
\paragraph{Web texts} On the web texts (SPR2), due to less previous works, we also use three label selection strategies as baselines: a \textit{majority} label baseline, a \textit{constant} strategy which always selects the positive label and a \textit{random} baseline which samples a positive target label according to the occurrence ratio in the training data (\textit{maj}, \textit{constant} \& \textit{ran}, Table \ref{tab:comprudteispr2}, left-hand side). 

Our basic \textbf{MarkerE} method yields massive improvements over both baselines (more than +10 pp.\ F1) in 4 out of 14 proto-role properties. For \textit{argument changes possession} and \textit{awareness} the improvement over both baselines is more than +25 pp.\ F1 and for \textit{sentient} more than +40 pp. However, in the \textit{partitive} property, the constant-label baseline remains unbeaten by a large margin (-21.7 pp.). Overall, all Marker models yield large performance increases over the baselines. For example, \textbf{MarkerE} yields significant improvements both over \textit{random} (+27.7 pp.\ macro F1), \textit{constant} (+9.5) and \textit{majority} (+45.0). 

Intriguingly -- while the contextual embeddings provide a massive performance boost on news texts (SPR1), -- they appear not to be useful for our model on the web texts. The macro F1 score of \textbf{MarkerEB} is slightly worse (-1 pp.) than that of \textbf{MarkerE} and the micro F1 score is only marginally better (+0.9). The same holds true for the single-model instances: \textbf{Marker} performs better than \textbf{MarkerB} by 2.2 pp.\ macro F1 albeit marginally worse micro F1 wise by 0.5 pp. 

Why exactly the contextual embeddings fail to provide any valuable information when labeling arguments in web texts, we cannot answer with certainty. A plausible cause could be overfitting problems stemming from the increased dimensionality of the input vectors. In fact, the contextual embeddings increase the number of word vector features by more than two times over the dimension of the GloVe embeddings. This inflates the number of parameters in the LSTM's weight matrices. As a consequence, the likelihood of overfitting is increased -- an issue which is further aggravated by the fact that SPR2 data are significantly fewer than SPR1 data. SPR2 contains less than five thousand predicate-argument training examples, roughly half the size of SPR1.

\label{par:spr2creation} Another source of problems may be rooted in the target-label construction process for SPR2. This question does not arise when using SPR1 data since all annotations were performed by a single annotator. The SPR2 data, in contrast, contains for each predicate-argument pair, two annotations. In total, the data was annotated by many crowd workers -- some of whom provided many and some provided few annotations. Perhaps, averaging Likert scale annotations of two random annotators is not the right way to transform SPR2 to a multi-label task. Future work may investigate new transformation strategies. For example, we can envision a strategy which finds reliable annotators and weighs the choices of those annotators higher than those of less reliable annotators. This should result in an improved SPR2 gold standard for both multi-label and multi-variate Likert scale SPRL systems.

\subsection{Likert Scale Regression Results}

\paragraph{News texts} \textbf{MarkerE} achieves large performance gains for the properties \textit{location} and \textit{change of location} ($\Delta \rho$: +0.136 \& $\Delta \rho$: +0.066, Table \ref{tab:comprudtei}). This is in accordance with the results for these two properties in the multi-label prediction setup. Our model is outperformed by RUD'18 
in the property \textit{stationary}
($\Delta\rho$: -0.045). 
All in all, \textbf{MarkerE} outperforms RUD'18 ($\Delta$ macro $\rho$: +0.005). When providing additional contextual word embeddings from the large language model, the correlations intensify for almost all role properties. Overall, the contextual embeddings in \textbf{MarkerEB} yield an observable improvement of +0.063 $\Delta$ macro $\rho$ over \textbf{MarkerE} (which solely uses GloVe embeddings).

\paragraph{Web texts} Our \textbf{MarkerE} model outperforms RUD'18 slightly by +0.001$\Delta\rho$ (Table \ref{tab:comprudteispr2}). 
 However, if we compare with RUD'18's model setup which achieved the best score on the SPR1 testing data (pre-training with a supervised MT task, macro regression result SPR2: 0.521$\rho$), we achieve a significantly higher macro-average ($\Delta\rho$: +0.014). Yet again, when our model is provided with contextual word embeddings, a large performance boost is the outcome. In fact, \textbf{MarkerEB} outperforms \textbf{MarkerE} by +0.045 $\Delta$ macro $\rho$ and RUD'18's best performing configuration by +0.046. This stands in contrast to the multi-labeling results on this type of data, where the contextual embeddings did not appear to provide any performance increase. As previously discussed (\S \ref{par:spr2creation}), this discrepancy may be rooted in the task generation process of SPR2 which requires transforming two annotations per example to one ground truth (the two annotations per example stem from two out of, in total, 50 workers). 

\subsection{Discussion}
\label{par:discuss}
Leaving the contextual word embedding inputs aside, the performance differences of our Marker models to RUD'18 may seem marginal for many properties. Albeit our \textbf{MarkerE} yields an observable improvement of 1.0 pp.\ macro F1 in the multi-label setup, in the regression setup the performance gains are very small ($\Delta \rho$ SPR1: +0.005, $\Delta \rho$ SPR2: +0.001, Table \ref{tab:comprudtei} \& \ref{tab:my_label}). In addition, our model as a single model instance (\textbf{Marker}) is outperformed by RUD'18's approach both in the regression and in the multi-label setup. However, it is important to note that the result of our system has substantially fewer dependencies (Table \ref{tab:my_label}). 

Firstly, our model does not rely on supplementary gold syntax -- in fact, since it is span-based, our model is completely agnostic to any syntax. Besides our approach, only REI'15 does not depend on supplementary gold syntax for the displayed results. However, \textit{all} of our models outperform REI'15's feature-based linear classifier in \textit{every} property (+17 pp.\ macro F1 in total) except for \textit{destroyed} (where \textbf{MarkerEB} performed slighly worse by -2.8 pp.\ F1). Also, results of SRL systems on semantic role labeling data show that span-based SRL systems often lag behind a few points in accuracy (cf.\ \citet{DBLP:journals/corr/abs-1901-05280}, Table 1). When provided with the syntactic head of the argument phrase, a model may immediately focus on what is important in the argument phrase. When solely fed the argument-span, which is potentially very long, the model has to find the most important parts of the phrase on its own and is more easily distracted. Additionally, identifying the head word of an argument may be more important than getting the boundaries of the span exactly right. In other words, span-based SPRL models may be more robust when confronted with slightly erroneous spans compared to dependency-based models which may be vulnerable to false positive heads. However, this hypothesis has to be confirmed or disproven experimentally in future work.

Further information about differences of various SPRL approaches is displayed in Table \ref{tab:my_label}.  In sum, despite having significantly fewer dependencies on external resources, our approach proves to be competitive with all methods from prior works, including neural systems. When combined in a simple ensemble, our model outperforms previous systems. When we feed additional contextual word embeddings, the  results can be boosted further by a large margin. In the following section, we show that ensembling SPRL models has another advantage besides predictive performance gains. Namely, it decreases SPRL model sensitivities towards different random initializations. As a result, we find that a simple neural voter committee (ensemble) offers more robust SPRL predictions compared to a randomly selected committee-member (single model).

\subsection{Analysis}
 
\begin{figure*}[ht]
\begin{subfigure}{0.5\textwidth}
\centering
\includegraphics[width=0.9\textwidth]{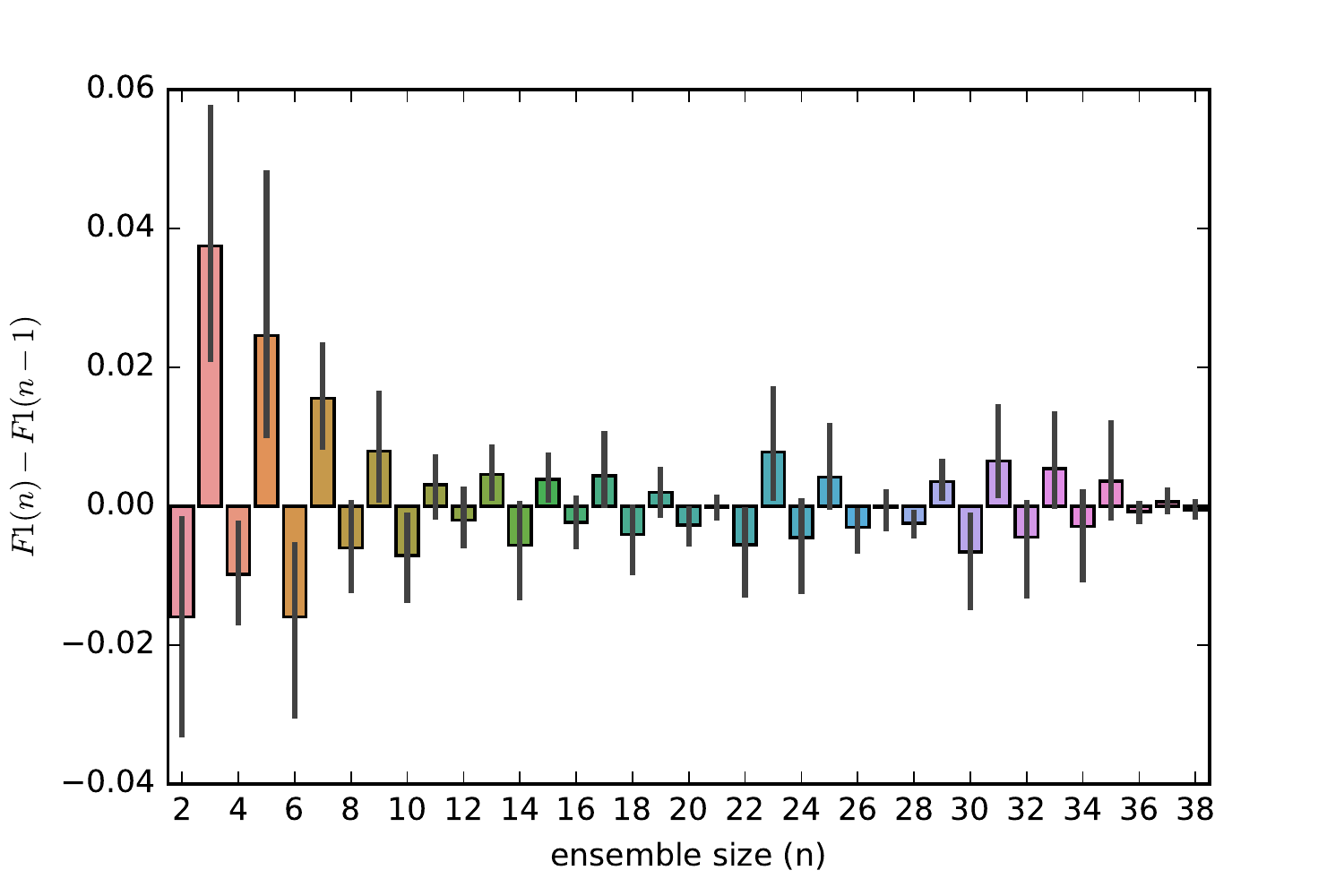}
\subcaption{Data: SPR1, voter: \textbf{Marker}.}

\end{subfigure}
\begin{subfigure}{0.5\textwidth}
\centering
\includegraphics[width=0.9\textwidth]{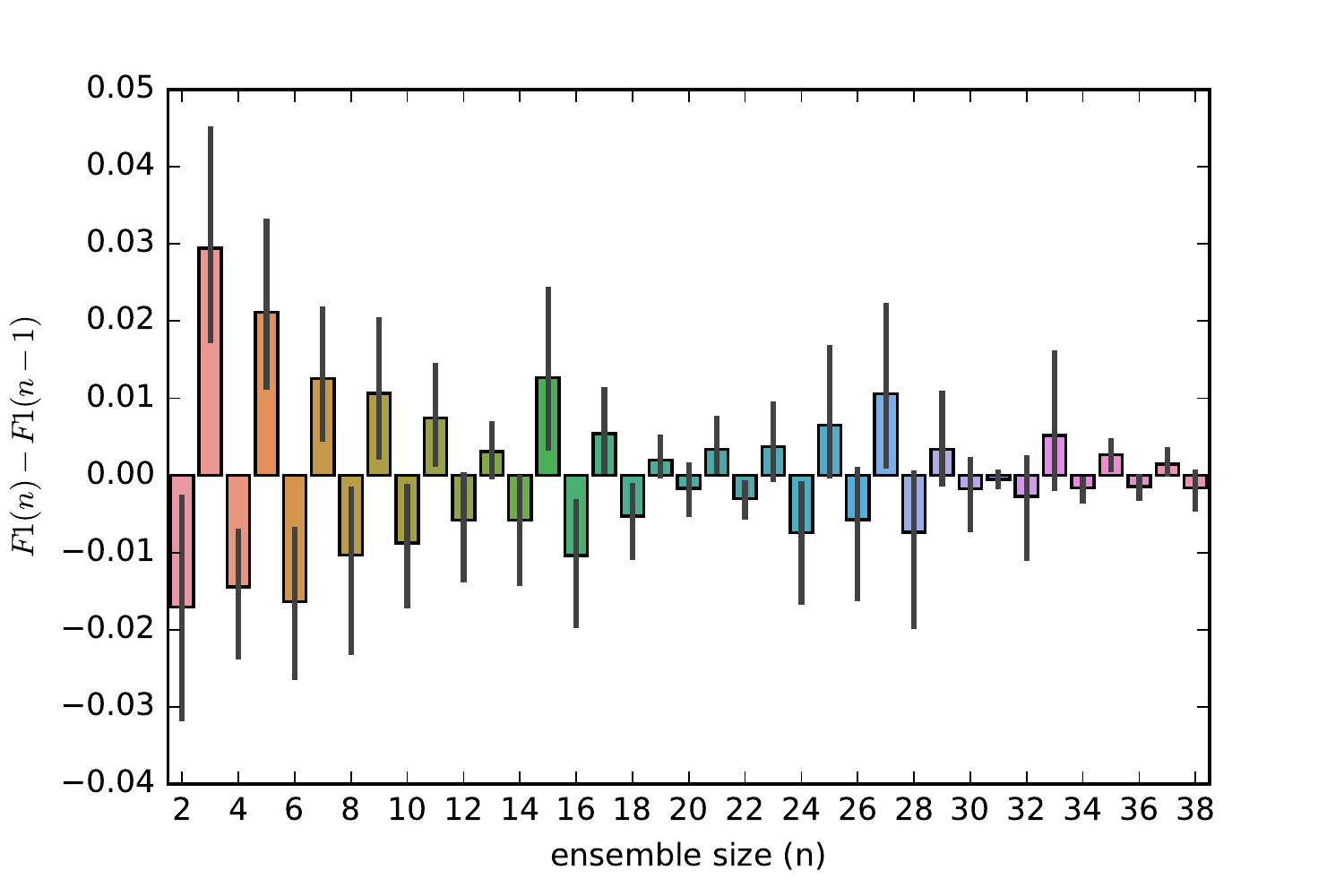}
\subcaption{Data: SPR1, voter: \textbf{MarkerB}.}
\end{subfigure}
\begin{subfigure}{0.5\textwidth}
\centering
\includegraphics[width=0.9\textwidth]{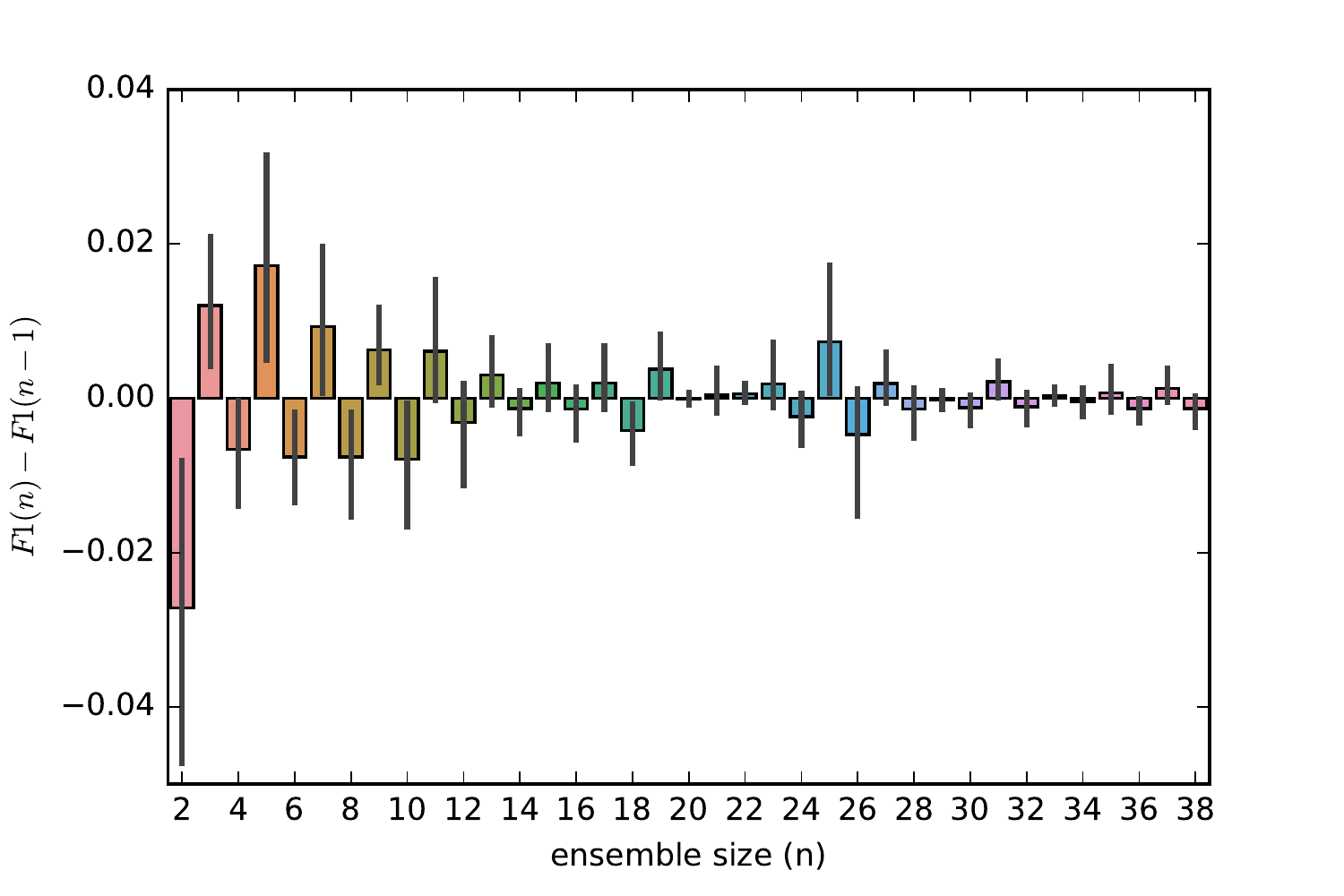}
\subcaption{Data: SPR2, voter: \textbf{Marker}.}

\end{subfigure}
\begin{subfigure}{0.5\textwidth}
\centering
\includegraphics[width=0.9\textwidth]{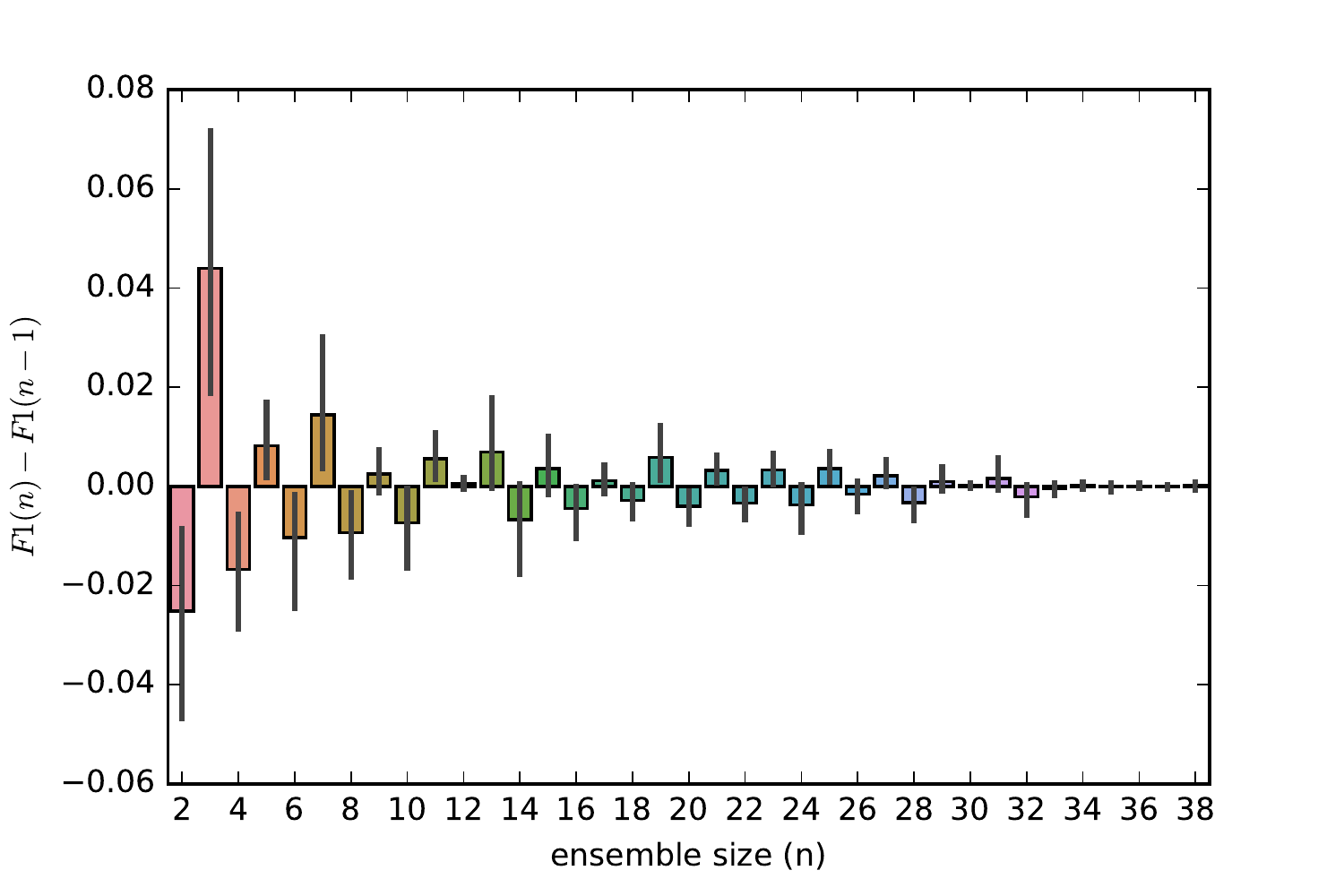}
\subcaption{Data: SPR2, voter: \textbf{MarkerB}.}
\end{subfigure}
\caption{Adding more voters leads to convergence in SPRL predictions. $x$-axis: number of voter models partaking in the ensemble. $y$-axis: F1 mean difference over all proto-roles from the ensemble with $x$ voters compared to the ensemble with $x-1$ voters. Thin bars represent standard deviations.}
\label{fig:difflast}
\end{figure*}
\paragraph{Ensembling increases accuracy} To investigate whether ensembling improves proto-role labeling results significantly, we conduct McNemar significance tests \citep{mcnemar1947note} comparing the predictions of \textbf{MarkerB} and \textbf{MarkerEB}.
\begin{table}
    \centering
    \scalebox{0.62}{\begin{tabular}{l|rr|rr|r}
    &\multicolumn{2}{c}{Ensembling worse}&\multicolumn{2}{c}{Ensembling better}\\
    \cmidrule{2-6}

            Data&$p\in [0.005,0.05)$&$p<0.005$ & $p\in [0.005,0.05)$&$p<0.005$& NS \\
            \midrule
         SPR1& 0 &1& 5 & 9 &3\\
         SPR2 & 1 &1 &5&2&5\\ 
         \bottomrule
    \end{tabular}}
    \caption{McNemar significance test results of \textbf{MarkerEB} against \textbf{MarkerB}. Counts of properties for which a significance category applies (NS: \#properties with insignificant difference).}
    \label{tab:significance}
\end{table}
The significance test results summarized in Table \ref{tab:significance} are unambiguous: for many proto-role properties, ensembling helps to improve performance significantly (SPR1: $\frac{14}{18}$ cases; SPR2: $\frac{7}{14}$ cases; significance level: $p<0.05$). However, for few cases, ensembling resulted in significantly worse predictions (SPR1: \textit{change of location}; SPR2: \textit{change of location}, \textit{instigated} and \textit{partitive}; significance level: $p<0.05$). For the rest of the properties, differences in predictions remain insignificant. 

\paragraph{Ensembling increases robustness} Additionally, we find that ensembling increases the robustness of our neural models. Consider Figure \ref{fig:difflast}, where we display the performance difference of a $n$-voter ensemble to the same ensemble after one additional voter joined ($n+1$-voter ensemble). The difference fluctuates wildly while the ensemble is still small. This suggests that a different random seed yields significantly different predictions. Hence, our a single neural Marker model is very vulnerable to the quirks of random numbers. However, when more voters join, we see that the predictions become notably more stable. An outcome which holds true for both data sets and two different ensemble model configurations (\textbf{MarkerE} and \textbf{MarkerEB}). We draw two conclusions from this experiment: (i) a single neural SPRL model is extremely sensitive to different random initializations. Finally, (ii) a simple voter ensemble has the potential to alleviate this issue. In fact, when we add more voters, the model converges towards stable predictions which are less influenced by the quirks of random numbers. 

\begin{table}
    \centering
    \scalebox{0.67}{\begin{tabular}{l|r|rrrrr|}
    
    &&\multicolumn{5}{c|}{ablated component}\\
    \cmidrule{2-7}
        Data & MarkerEB & SelfAtt&mark.\ &pred-mark.\ &arg-mark.\ &hier.\ \\
        \midrule
         SPR1 & 77.5 &73.1 &50.7& 76.3 &60.1& 76.7 \\
         SPR2 & 69.9 &67.3 &59.3&68.2&63.3& 68.9\\ 
         \bottomrule
    \end{tabular}}
    \caption{Multi-labeling F1 macro scores for different MarkerEB model configurations over SPR1 and SPR2.}
    \label{tab:ablations}
\end{table}

\paragraph{Model Ablations} All ablation experiments are conducted with \textbf{MarkerEB} in the multi-label formulation. We proceed by ablating different components in a leave-one-out fashion: (i) the self-attention components of the ensemble model are removed (SelfAtt in Table \ref{tab:ablations}); (ii) we abstain from highlighting (a) the arguments and predicates, (b) only the predicates and (c) only the arguments (mark.\,pred-mark.\ and arg-mark.\ in Table \ref{tab:ablations}); Finally, (iii) we remove the hierachical structure and do not predict auxiliary outputs (hier.\ in Table \ref{tab:ablations}).

From all ablated components, removing simultaneously both predicate and argument-markers hurts the model the most (SPR1: -26.8 pp.\ macro F1; SPR2: -10.6). Only ablating the argument-marker also causes a great performance loss (SPR1: -17.4, SPR2: - 6.6). On the other hand, when only the predicate marker is ablated, the performance decreases only slightly (SPR1: -1.2, SPR2: -1.7). In other words, it appears to be of paramount importance to provide our model with indicators for the argument position in the sentence, but it is of lesser importance to point at the predicate. The self-attention component can boost the model's performance by up to +4.4  pp.\ F1 on SPR1 and +2.6 on SPR2. The hierarchical structure with intermediate auxiliary Likert scale outputs leads to gains of approximately +1 pp.\ macro F1 in both data sets. This indicates that indeed the finer Likert scale annotations provide auxiliary information of value when predicting the labels at the top layer, albeit the performance difference appears to be rather small.

\section{Conclusion}

In our proposed SPRL ensemble model, predicate-argument constructs are highlighted with concurrently learned marker embeddings and self-attention enables the model to capture long-distance relationships between arguments and predicates. 
The span-based method is competitive with the dependency-based state-of-the-art which uses gold heads. When combined in a simple ensemble, the method overall outperforms the state-of-the-art on newspaper texts (multi-label prediction macro F1: +1.0 pp.). When fed with contextual word embeddings extracted from a large language model, the method outperforms the state-of-the-art by 6.4 pp.\ macro F1. Our  method is competitive with the state-of-the-art for Likert regression on texts from the web domain. In the multi-label setting, it outperforms all baselines by a large margin. Furthermore, we have shown that a simple Marker model voter ensemble is very suited for conducting SPRL, for two reasons: (i) results for almost every proto-role property are significantly improved and (ii) considerably more robust SPRL predictions are obtained. 

We hope that our work sparks more research into semantic proto-role labeling and corpus creation. Dowty's feature-based view on roles allows us to analyze predicate-argument configurations in great detail -- an issue which we think is located in the marrow of computational semantics. 

\section*{Acknowledgements}

This work has been supported by the German Research Foundation as part of the Research Training Group Adaptive Preparation of Information from Heterogeneous  Sources  (AIPHES)  under  grant no.\ GRK 1994/1 and by  the  Leibniz  ScienceCampus ``Empirical  Linguistics  and  Computational  Language  Modeling'', supported by the Leibniz Association under grant no.\  SAS-2015-IDS-LWC  and  by  the  Ministry  of  Science, Research, and Art of Baden-W\"urttemberg.

\bibliography{proto_ensemble}
\bibliographystyle{acl_natbib}

\appendix

\section{Supplemental Material}
\label{sec:supplemental}
\subsection{Notes}
\label{sec:notes}

\paragraph{Calculation of macro F1} The global performance metric for multi-label SPRL is defined as `macro F1'. To ensure full comparability of results, we use the same formula as prior works \cite{DBLP:journals/corr/abs-1804-07976}: 
\begin{equation}
   \frac{2 \cdot P_{macro~avg.} \cdot R_{macro~avg.}}{P_{macro~avg.} + R_{macro~avg.}},
\end{equation}
where $P$ and $R$ are \textit{Precision} and \textit{Recall} and \textit{macro avg.} means the unweighted mean of these quantities computed over all proto-role properties. 
The above macro F1 metric, though not explicitly displayed in the prior work papers, has been confirmed by the main authors (email).

\paragraph{Calculation of Pearson's $\rho$} Person's $\rho$ quantifies the linear relationship between two random variables $X$ and $Y$. Computed over a sample $\{(x_i,y_i)\}_1^n$ it is calculated with the following formula:
\begin{equation}
   \rho =\frac{\sum ^n _{i=1}(x_i - \bar{x})(y_i - \bar{y})}{\sqrt{\sum ^n _{i=1}(x_i - \bar{x})^2} \sqrt{\sum ^n _{i=1}(y_i - \bar{y})^2}}
\end{equation}

Given $|P|$ proto-role properties and corresponding correlation coefficients $\rho_1,...\rho_{|P|}$, the macro Pearson's $\rho$ is calculated as

\begin{equation}
    macro~\rho=\frac{\sum_{i=1}^{|P|}\rho_i}{|P|}.
\end{equation}

\paragraph{Data split of SPR1} \cite{DBLP:conf/aaai/TeichertPDG17} re-framed the SPRL task as a multi-label problem. Previously the task was to answer, given a predicate and an argument, one specific proto-role question (binary label or single output regression). Now we need to predict all proto-role questions at once (multi-label or multi-ouput regression). In order to allow this formulation of the task, the authors needed to redefine the original train-dev-test split of SPR1 (recent works, including ours, all use the re-defined split).

\paragraph{Reported Numbers} 
In the EMNLP publication of \citet{DBLP:journals/corr/abs-1804-07976} we found a few minor transcription errors in the result tables (confirmed by email communication with the main authors, who plan to upload an errata section). In the case of transcription errors, we took the error-corrected numbers which were sent to us via email. 

\subsection{Hyperparameters \& Preprocessing}
\label{sec:hyp}
\begin{table}
    \centering
    \scalebox{0.69}{\begin{tabular}{l|l}
    \toprule
    hyper-parameter &choice\\
    \midrule
       $\lambda$ (main loss)  & 1 \\
        $\lambda'$ (aux.\ losses)  & 0.2\\
        optimizer & Adam \citep{DBLP:journals/corr/KingmaB14}\\
        optimizer param.\ &$\beta_1=0.9,\beta_2=0.999,\epsilon=1e^{-07}$\\
        learning rate &0.001\\
        Bi-LSTM units & $2 \cdot 64$\\
        max.\ seq length & 30 \\
        padding & pre \\
        Marker embeddings,init  & $U(-0.05,+0.05)$\\
        Fixed embeddings, init & GloVe 300d \cite{pennington2014glove}\\
        
    \bottomrule
    \end{tabular}}
    \caption{Hyper parameter configuration.}
    \label{tab:hparams}
\end{table}

The hyper parameter configuration of our model are displayed in Table \ref{tab:hparams}. \textit{Sequence pruning}: Consider that $I=\{i\}$ is the index of the predicate and $J=\{j,...,k\}$ are the indices corresponding to the argument. As long as the input sequence length is longer than maximum length (30, cf. Table \ref{tab:hparams}), we clip left tokens so that the index of the token $m < \min{I \cup J}$, then we proceed to clip tokens to the right so that $m > \max{I \cup J}$, for the very rare cases that this was not sufficient we proceed to clip tokens with $m \notin I \cup J$ (the marker sequences are adjusted accordingly). The clipping strategy ensures that predicate and argument tokens are present in every input sequence. Sequences which are shorter than 30 words are \textit{pre}-padded with zero vectors. 

\end{document}